\documentclass[alpha-refs]{wiley-article}

\usepackage{graphicx}
\usepackage{float}
\usepackage[space]{grffile}
\usepackage{latexsym}
\usepackage{textcomp}
\usepackage{longtable}
\usepackage{tabulary}
\usepackage{booktabs,array,multirow}
\usepackage{amsfonts,amsmath,amssymb}
\usepackage{color,soul}
\usepackage{todonotes}
\usepackage{natbib}
\usepackage{url}
\usepackage{hyperref}
\hypersetup{colorlinks=false, pdfborder={0 0 0}}
\usepackage{etoolbox}
\usepackage{amsmath}
\usepackage{svg}
\usepackage[]{algorithm2e}
\usepackage[table]{xcolor}
\makeatletter

\newcommand{\indep}{\perp \!\!\! \perp}

\sethlcolor{orange}

\makeatother
\newif\iflatexml\latexmlfalse

\AtBeginDocument{\DeclareGraphicsExtensions{.pdf,.PDF,.eps,.EPS,.png,.PNG,.tif,.TIF,.jpg,.JPG,.jpeg,.JPEG}}

\usepackage[utf8]{inputenc}
\usepackage[english]{babel}

\usepackage{siunitx}

\iflatexml

\else

\paperfield{}

\abbrevs{TIMSS: Trends in International Mathematics and Science Study, 
BART: Bayesian Additive Regression Trees, 
BCF: Bayesian Causal Forests}

\corraddress{Nathan McJames, \\Hamilton Institute and Department of Mathematics and Statistics, \\Maynooth University, Maynooth, \\Co. Kildare, Ireland.}

\corremail{nathan.mcjames.2016@mumail.ie}

\papertype{Original Article}

\title{Bayesian Causal Forests for Multivariate Outcomes: Application to Irish Data From an International Large Scale Education Assessment.}

\author[1, 2]{Nathan McJames}
\author[1, 2]{Andrew Parnell}
\author[1, 2]{Yong Chen Goh}
\author[2]{Ann O'Shea}

\affil[1]{Hamilton Institute, Maynooth University}
\affil[2]{Department of Mathematics and Statistics, Maynooth University}

\runningauthor{McJames et al.}

\begin{document}

\maketitle

\selectlanguage{english}

\begin{abstract}
Bayesian Causal Forests (BCF) is a causal inference machine learning model based on a highly flexible non-parametric regression and classification tool called Bayesian Additive Regression Trees (BART). Motivated by data from the Trends in International Mathematics and Science Study (TIMSS), which includes data on student achievement in both mathematics and science, we present a multivariate extension of the BCF algorithm. With the help of simulation studies we show that our approach can accurately estimate causal effects for multiple outcomes subject to the same treatment. We also apply our model to Irish data from TIMSS 2019. Our findings reveal the positive effects of having access to a study desk at home (Mathematics ATE 95\% CI: [0.20, 11.67]) while also highlighting the negative consequences of students often feeling hungry at school (Mathematics ATE 95\% CI: [-11.15, -2.78] , Science ATE 95\% CI: [-10.82, -1.72]) or often being absent (Mathematics ATE 95\% CI: [-12.47, -1.55]).

\end{abstract}%

\section{Introduction}

Estimating the causal effect of an intervention on an outcome variable of interest is an important but difficult task. When attempting to do this, the key challenge we are presented with is that of disentangling correlations from any direct causal influence that may be present. This issue is ideally circumvented by designing experiments in which assignment to the treatment group is completely randomised \citep{sibbald1998understanding}. Such studies are often impossible, however, due to financial, ethical, or other considerations \citep{west2008alternatives}. For this reason, it is often necessary for researchers to conduct their analyses on observational data in which assignment to the treatment group cannot be guaranteed to be random. A further complication occurs when working with observational data as there may be confounding variables present which have an impact on both the probability of being assigned to the treatment group and the outcome of interest. Confounding variables can bias the results of analyses by making it appear as if there is a direct causal relationship between two variables when in fact this is not the case \citep{greenland1999confounding}. Special approaches which attempt to remove this biasing effect are therefore necessary when working with observational data.

Common approaches to address this problem within the causal inference literature include matching, propensity score methods, and Bayesian causal networks \citep[see][for a review]{yao2021survey}. Matching methods attempt to replicate the conditions that would be present in a randomised controlled trial by balancing the distribution of covariates among the treated and controlled units \citep{stuart2010matching}. This balance can be achieved in a number of different ways but invariably involves using some measure of similarity to match individuals with similar characteristics in the control and treatment groups. Propensity score methods encompass a variety of widely-used techniques which estimate the probability of being assigned to an intervention \citep{pan2018propensity}. Propensity score weighting is one such example which creates balance across control and treatment groups by weighting observations according to their propensity score \citep{kurz2022augmented}. Finally, popularised by \citet{pearl1995bayesian}, causal graphs and Bayesian causal networks allow the identification of causal effects by using directed acyclic graphs to represent causal relationships between observed variables. Recently there has been a surge of interest in applying advanced machine learning models in this area \citep{caron2020estimating}. Our approach (described below) is best categorised as belonging to this final group of machine learning based methods.

The machine learning family of methods includes a vast assortment of meta-learner techniques which exploit the predictive capabilities of machine learning models in order to estimate heterogeneous treatment effects. A key strength of this family of methods is that it is very flexible, and many of the the relevant techniques can be performed with virtually any machine learning model. One of the most important contributions to this area was made by \citet{hill2011bayesian} who demonstrated that by using a sufficiently flexible regression model such as Bayesian Additive Regression Trees \citep[BART,][]{chipman2010bart}, it is possible to accurately estimate treatment effects. A second influential contribution was made by \citet{hahn2020bayesian} who built on Hill's work by using \citeauthor{robinson1988root}'s (\citeyear{robinson1988root}) treatment effect parameterisation to separate the estimation of $Y$ into a prognostic effect $\mu$, and a treatment effect $\tau$. This approach, named Bayesian Causal Forests (BCF), has a number of advantages over that of Hill as it allows separate priors to be applied to the $\mu$ and $\tau$ components of the model, and enables individual level treatment effects to be estimated directly from the data.

An important limitation of many causal machine learning methods, including Bayesian Causal Forests, is that they are only applicable to a single outcome variable subject to a binary treatment $Z$. Therefore, motivated by data from the Trends in International Mathematics and Science Study \citep{mullis2020timss}, which includes data on both the mathematics and science achievement of eighth grade (approximately 14 - 15 year old) secondary school students, we present a multivariate extension of BCF which is capable of estimating the causal effect of an intervention on multiple outcomes simultaneously. With our new approach, we consider the effect of a number  of home-related factors on student achievement. Specifically, we attempt to answer the following three research questions: 1) "What effect does having access to a study desk at home have on student achievement in mathematics and science?", 2) "What is the impact on student achievement in mathematics and science of often arriving at school feeling hungry?", and 3) "What effect does regular absence from school have on student achievement in mathematics and science?". We investigate these factors because they have important implications for student focused initiatives such as free school meals programmes and back to school allowances which are designed to assist students from disadvantaged backgrounds \citep{taras2005nutrition, kennedy2013key}.

The main advantage of our multivariate approach is a potentially substantial reduction in the uncertainty associated with the causal parameters, since the model now has access to extra information through the other treatment variables. Our approach shares similarities with that of a recent multivariate extension of Bayesian Factor Analysis models for causal inference which demonstrates the potential for a multivariate approach to causal inference \citep{samartsidis2020bayesian}. Our work also shares similarities with that of \citet{segal2011multivariate} who developed a multivariate extension of random forests, and demonstrated the potential for multivariate tree based models. Our work is different from these two studies because, first in the case of \citeauthor{segal2011multivariate}, our focus is not on modelling the outcome variables themselves, but instead we are interested in discovering how they respond to a given treatment. Second, the multivariate causal Factor Analysis model developed by \citeauthor{samartsidis2020bayesian} uses a very different structure to our BART based model. We believe our approach offers greater flexibility and may be used in a much wider variety of settings.

The remainder of our paper is organised as follows: In Section~\ref{TIMSS Section} we give some background on the Trends in International Mathematics and Science Study, the dataset motivating our multivariate approach. Section~\ref{BART Description} describes Bayesian Additive Regression Trees, the model providing the foundation upon which Bayesian Causal Forests are built. Section~\ref{BCF Description} explains how BCF leverages the impressive predictive capabilities of BART for the purpose of estimating heterogeneous treatment effects, and Section~\ref{MVBCF Description} details the modifications necessary to extend BCF to the multivariate setting. In Section~\ref{Simulation Section} we present the results of a simulation study, in which we demonstrate the substantial benefits of jointly modelling all outcome variables available. In Section~\ref{Application Section} we apply our multivariate extension of BCF to the motivating dataset, TIMSS 2019. Here, we investigate the effects of a number of treatments on student mathematics and science achievement, including home study supports, being hungry at school, and absenteeism. We conclude our paper with a discussion of our results, the limitations, and potential avenues for future research.

\section{Trends in International Mathematics and Science Study}
\label{TIMSS Section}

The Trends in International Mathematics and Science Study (TIMSS) is a large scale international study organised by the International Association for the Evaluation of Educational Achievement (IEA). It has taken place in many countries across the world every four years since 1995, with 64 countries participating in TIMSS 2019. As part of the study, students in the fourth and eighth grade of secondary school (typically aged approximately 10 - 11 and 14 - 15 respectively) are given a short assessment in mathematics and science, which is used to estimate their overall achievement level. The eighth grade students also complete a short background questionnaire on topics such as their home and classroom environment, and how much they like and feel confident in these subjects. The teachers and principals of these students are also given short questionnaires on their educational background, teaching practices, and school access to learning resources, thus providing us with a large number of covariates to control for as potential confounding variables. This makes TIMSS an excellent source of information for researchers investigating factors associated with student confidence and achievement in mathematics and science.

Due to its scale and comprehensive nature, TIMSS data has been the subject of many studies in the field of education since its origin in 1995. Some recent studies using data from TIMSS include \citet{tang2022impact} who investigate the impact of science teacher continual professional development on student achievement in science, and \citet{chen2022effects} who considers the effect of the interaction between classroom and individual achievement levels on student confidence in mathematics. Our focus in this paper however will be on the effect of three specific treatments on student achievement in mathematics and science. In contrast to much of the existing literature which focuses on typically just one of these outcomes, we will model achievement in both subjects jointly.

The causal factors we will focus on in this study are related to a student's household environment: home study supports, often feeling hungry at school, and absenteeism. Home study supports and educational resources are known to be important predictors of student achievement \citep{tsai2015hierarchical}. This finding has been consistent across multiple different studies, but given that home educational resources are likely to be linked to socioeconomic status, it is important to ascertain the true nature of this relationship within a causal framework. The negative consequences of students lacking the opportunity to have a healthy breakfast in the morning are also well documented \citep{vik2022aspects}. Hungry students can find it difficult to concentrate in class and can be disadvantaged by not meeting their daily nutritional requirements. As a result, free school meal programmes are sometimes provided to ensure all students can be ready for learning throughout the day \citep{taras2005nutrition}. Being hungry at school may be linked to a student's socioeconomic background and this must be accounted for as a source of confounding. 

The final factor under investigation in this paper is absenteeism, which has been linked to lower achievement levels \citep{vesic2021role}. As before, there is need for caution.  Research has found, for example, that bullying can be a major contributory factor to higher rates of absenteeism \citep{bennour2021multilevel}. However bullying itself is known to lead to lower levels of achievement \citep{ladd2017peer}, so this is an example of a confounding factor. This highlights the need to account for other important influences on achievement when investigating the relationship between two variables. Of course, there can also be many other reasons for students missing school such as illness and unapproved absences, but in this study we will not be able to make careful distinctions between them, as we are not provided with this information in the student questionnaire.

In summary, TIMSS is an excellent source of information for researchers in the field of education. TIMSS data has been used extensively to answer many important research questions over the years, but as the short discussion above highlights, there can often be multiple layers of complexity with the potential to bias the estimates of these analyses. Furthermore, much of the existing research has focused solely on achievement in one subject, employing traditional approaches such as multiple linear regression models which are not well suited to answering questions of a causal nature. For this reason we propose that a multivariate causal approach, capable of flexibly accounting for the many confounding variables that may be present, is well suited to these data. 

\section{Bayesian Non-parametric Estimation of Heterogeneous Treatment Effects}
\label{Model Section}

One of the fastest growing areas of research in the causal inference machine learning literature is the application of Bayesian non-parametric machine learning algorithms for the estimation of heterogeneous treatment effects. Before discussing these approaches in detail, however, we must first cover some notation. In this paper we will adopt the Neyman-Rubin causal model \citep{splawa1990application, rubin1974estimating, sekhon2008neyman} which can be applied to situations where we are interested in the effect of a treatment $Z$ on an outcome $Y$. The Neyman-Rubin causal model is based on the concept of potential outcomes, which asserts that for each observation $i$, there are two potential outcomes: one that would be observed under treatment $y_{i}(Z_i=1)$, and one that would be observed under control, $y_{i}(Z_i=0)$. Knowing both $y_{i}(Z_i=0)$ and $y_{i}(Z_i=1)$ would allow us to calculate the individual treatment effect for unit $i$, $\tau_{i}=y_{i}(Z_i=1)-y_{i}(Z_i=0)$. This is of course impossible, because we only ever observe one of the potential outcomes, and this is known as the fundamental problem of causal inference.

Although we may not observe both potential outcomes directly, we can estimate them with $\hat{y}_{i}(Z_i=0)$ and $\hat{y}_{i}(Z_i=1)$. Then, in the presence of the correct conditions, we may estimate $\tau_{i}$ with $\hat{\tau}_{i} = \hat{y}_{i}(Z_i=1) - \hat{y}_{i}(Z_i=0)$. The conditions which are required to hold for the reliability of this approach are, from \citet{kurz2022augmented}:

\begin{enumerate}
    \item The stable unit treatment value assumption (SUTVA). This requires that the potential outcomes of any individual $i$ must not be affected by the treatment status of any other individual $j$. For example, if student $j$ is often absent from school, this must not influence the achievement level of any other student $i$.
    \item The ignorability assumption. Also known as the unconfoundedness assumption, we require that there must be no confounding variables we cannot control for, or that were not collected as part of the study: $y_{i}(Z_i=1), y_{i}(Z_i=0) \indep Z_{i}|x_{i}$. 
    \item The overlap assumption. This requires that the propensity score for any individual $i$ must be bounded away from zero and one: $0<P(Z_i=1|x_{i})<1$. For example, if it was true that students from disadvantaged backgrounds were guaranteed never to have a study desk, this would be a violation of the overlap assumption.
\end{enumerate} 

The above provides us with a very flexible approach for estimating individual treatment effects, as $y_{i}(Z_i=1)$ and $y_{i}(Z_i=0)$ can be estimated with any sufficiently accurate model, $f$. A good choice for $f$, for a number of reasons, is Bayesian Additive Regression Trees, and this is what we will discuss next.

\subsection{Bayesian Additive Regression Trees}
\label{BART Description}

Bayesian Additive Regression Trees (BART) is a Bayesian non-parametric machine learning algorithm that is well suited to a variety of regression and classification tasks \citep{chipman2010bart}. BART can be described as a tree based ensemble method for predicting an unknown function $f(X)$ based on the contributions of many shallow trees. Individually these trees act as weak learners, each only explaining a small part of the unknown function, but when combined they are able to capture very complicated relationships and interactions between variables in the data. Owing to its impressive predictive performance, BART has become popular with researchers from many disciplines and has been used for a diverse range of applications in many fields such as medicine, economics, and education \citep{pierdzioch2016precious, sparapani2016nonparametric, mcjames2022factors}. BART is a very flexible model, which has enabled researchers to adapt or modify the underlying algorithm for various specialised use cases such as genomics, problems with local linearities and, of course, causal inference \citep{sarti2022bayesian, prado2021bayesian, hill2020bayesian}. 

Given an outcome variable $y$ of length $n$, and a covariate matrix $X$ consisting of $n$ observations of $d$ variables, the BART model can be written as follows:

\[y_{i}=\sum_{j=1}^{J}g(T_{j}, M_{j}, x_{i})+\epsilon_{i},\ \ \epsilon_{i}\sim N(0, \sigma^2)\]
\noindent
where $g()$ is a function which calculates the individual contribution of each tree $j$ of $J$ total trees. $M_{j}$ specifies the terminal node parameters associated with $j^{th}$ tree $T_{j}$. The residuals, $\epsilon_{i}$, are assumed to be normally distributed with mean $0$ and variance $\sigma^2$. Being a Bayesian model, appropriate priors are required for $T_{j}$, $M_{j}$ and $\sigma^2$.

The BART model is fitted using Markov Chain Monte Carlo with a combination of Gibbs sampling and Metropolis Hastings steps. The structure of the $j^{th}$ tree is updated at each iteration by choosing at random one of four possible operations to propose a new updated tree; grow, prune, change, or swap. If grow is selected, then a splitting rule is assigned to a randomly chosen terminal node which then becomes the parent of two children. If prune is selected, then a parent of two terminal nodes is chosen at random, and its children are removed from the tree. During the change operation, an internal node is chosen at random and its splitting rule is replaced with a new randomly chosen split rule. Finally, the swap operation selects a parent-child pair which are both internal nodes, and swaps their splitting rules with each other. 

To prevent any member of the ensemble from growing too large, a prior $P(T_{j})$ is placed on the structure of the $j^{th}$ tree. This prior specifies that the probability of any node at depth $d$ being non-terminal is given by $\alpha(1+d)^{-\beta}$. Therefore, for a tree $T_{j}$ with terminal nodes $h_{1} ... h_{K}$, and non-terminal nodes $b_{1} ... b_{L}$, we have that:

\[P(T_{j})=\prod_{k=1}^{K} \alpha(1+d(h_{i}))^{-\beta} \prod_{l=1}^{L} [1-\alpha(1+d(b_{l}))^{-\beta}]\]
\noindent
where $d()$ is a function for returning the depth of an arbitrary node, and $\alpha$ and $\beta$ are hyper parameters which can be adjusted to place a higher probability on the preferred tree depth. The purpose of this prior is to encourage more shallow trees, thus restricting the amount of variance any one tree can explain, and helping to avoid overfitting.

\begin{figure}
    \centering
    \includegraphics[height=5cm]{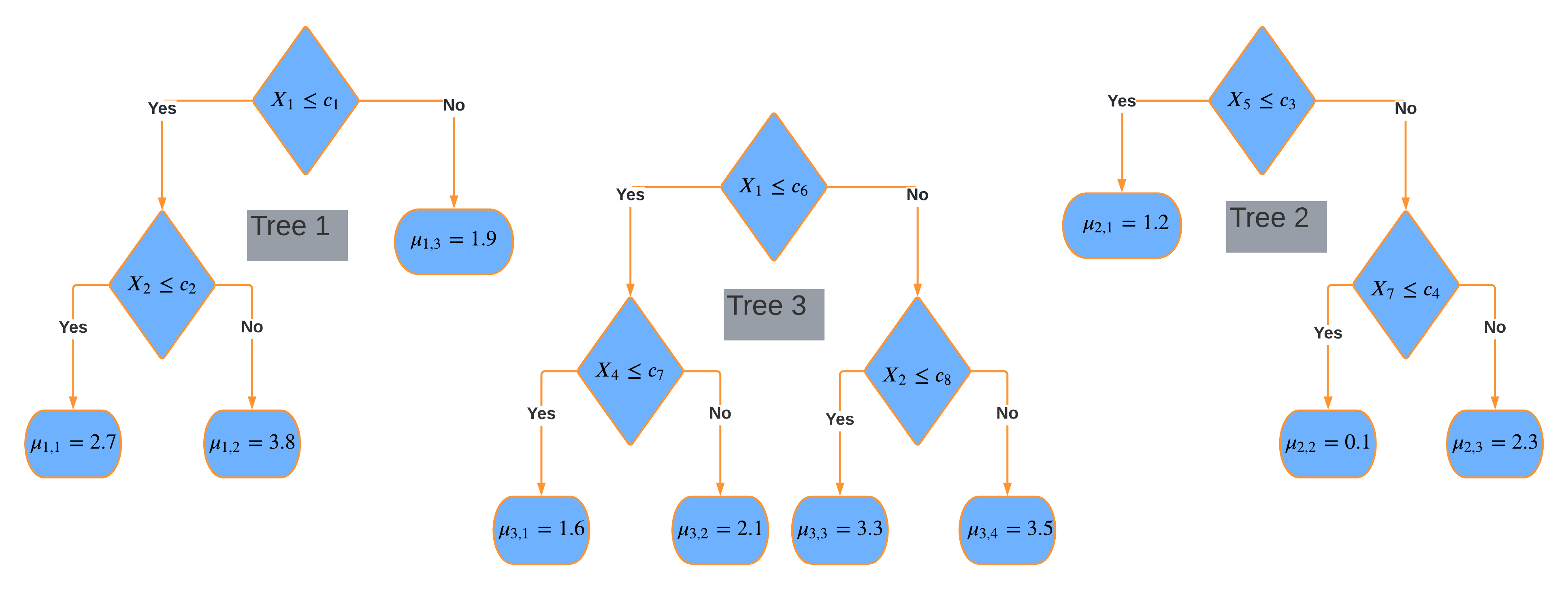}
    \caption{\label{ExampleTree}Diagram of a BART model with three decision trees as part of the ensemble. The predictions for an observation are given by following the decision rules from the root to the terminal nodes of the trees and summing the individual contributions together. For example, for an observation with $X_{1}>c_{1}$, $X_{5}<c_{3}$, $X_{1}<c_{6}$ and $X_{4}<c_{7}$, the final prediction would be given by $1.9+1.2+1.6=4.7$.}
\end{figure}

With the structure of the $j^{th}$ tree defined, the decision rules at each node form a pathway directing observations to the leaves of the tree. See Figure~\ref{ExampleTree} for an example. Terminal node parameters $\mu_{j,k}$ are now assigned to each of the $K$ leaves of the $j^{th}$ tree, responsible for providing a small but important contribution to the final prediction made by the model. To safeguard against any individual trees becoming unduly influential in $g()$, and to ensure that the scale of the $\mu$ parameters is sufficient to cover the whole of the observed data, the prior $\mu_{j,k} \sim N(0, \sigma_{\mu}^2)$ is used. With $y$ scaled during data pre-processing to follow a standard normal distribution, a sensible choice for the hyper parameter $\sigma_{\mu}^2$ is $1/J$, which places a high prior probability over the range of all observed $y$ values. The combination of priors above allows the likelihood used in the Metropolis-Hastings step to be calculated in closed form as a multivariate normal distribution summed across terminal nodes. The data enter this multivariate normal distribution as partial residuals calculated from the response minus the predictions of the other trees that are not being updated. 

When the terminal node parameters for tree $T_{j}$ have all been sampled, a newly selected grow/prune/change/swap operation is applied to tree $T_{j+1}$ and the process is repeated until all $J$ trees in the ensemble have had their terminal node parameters updated. At the end of each iteration the combined contribution from all $J$ trees is calculated and the precision parameter $1/\sigma^2$ is drawn from a Gamma posterior distribution which is conjugate to $1/\sigma^2 \sim Ga \left( \frac{\nu}{2}, \frac{\nu \lambda}{2} \right)$, where $\nu$ and $\lambda$ are the prior shape and rate hyper parameters. The above process repeats for a pre-specified number of iterations and the end result is a posterior distribution of trees, and terminal node and $\sigma^2$ parameters. The entire procedure can be summarised as in Algorithm~\ref{BARTAlgorithm}.

\RestyleAlgo{boxruled}
\begin{algorithm}
 \KwData{Target variable $y$ of length $n$ (Mean 0, SD 1), Covariate matrix $X$ ($n$ rows and $d$ columns)}
 \KwResult{Posterior of trees $T$, Samples of $\sigma$, Samples of $\hat{y}$}
 \textbf{Initialisation\;}
 	Set Hyper-parameter values of $\alpha$, $\beta$, $\sigma_\mu$, $\nu$, $\lambda$\;
	Set Number of trees $J$, Number of iterations $N_{iter}$\;
	Set Initial value $\sigma^2 = 1$, and set trees $T_j$; $j=1, \ldots, M$ to stumps with terminal node parameters set to 0\;
\For{iterations $i$ from 1 to $N_{iter}$}{
	\For{trees $j$ from 1 to $J$}{
		Compute partial residuals $R_{j}$ from $y$ minus predictions of all trees except tree $j$\;
		Grow a new tree $T_j^{new}$ based on grow/prune/change/swap\;
		Accept/Reject newly proposed tree structure with Metropolis-Hastings step using $P(T_{j}|R_{j}, \sigma)\propto P(T_{j})P(R_{j}|T_{j}, \sigma)$\;
		Sample $\mu$ values using $P(M_{j}|T_{j}, R_{j}, \sigma)$\;
	}
	Combine predictions from all trees to get $\hat{y}$\;
	Update $\frac{1}{\sigma^2}$ using $P(\frac{1}{\sigma^2} | \hat{y})$\;
}
\caption{\label{BARTAlgorithm}Bayesian Backfitting MCMC Algorithm for BART}
\end{algorithm}

\subsection{Bayesian Causal Forests}
\label{BCF Description}

Bayesian Causal Forests is an advanced causal inference machine learning algorithm \citep{hahn2020bayesian}. BCF uses BART as a foundation for estimating causal effects and shares the same desirable features such as impressive predictive performance, careful regularisation through the use of Bayesian priors, and uncertainty quantification. BCF does however have a number of advantages over BART for estimating heterogeneous treatment effects, and this is made possible by adopting the Robinson parameterisation which expresses the outcome $y$ as:

\[y_{i}=\mu(x_{i}, \hat{\pi}_{i})+\tau(x_{i})Z_{i}+\epsilon_{i}\]
\noindent
where $\mu()$ and $\tau()$ are both BART ensembles which work together to estimate two separate parts of $y$: a prognostic effect $\mu$, which can be thought of as the expected outcome under control when the treatment variable $Z$ is coded as 1 for treatment, 0 for control, and a treatment effect $\tau$, which can be interpreted as the impact on $y$ of receiving treatment. The additional covariate $\hat{\pi}_{i}$ included in the $\mu()$ part of the model is the propensity score described earlier, which is simply the estimated probability of individual $i$ receiving treatment: $\hat{\pi}_{i} = P(Z_{i} = 1)$. The inclusion of the propensity score in $\mu()$ is important for avoiding a phenomenon called regularisation induced confounding, and is especially useful in situations where the likelihood of receiving treatment is in some way related to the expected outcome under control \citep{hahn2020bayesian}. We will therefore include an estimate of $\hat{\pi}_{i}$, obtained using a BART model, in all experiments in this study.

The parameterisation above has a number of important benefits associated with it. First, it allows different amounts of regularisation to be applied to the $\mu$ and $\tau$ parts of $y$. It is common to apply greater regularisation to $\tau$ than to $\mu$ because we expect the degree of heterogeneity in the treatment effects to be relatively simple in comparison to $y$ itself. This prior belief can be incorporated into the model by using a smaller $\alpha$ and larger $\beta$ in the prior $P(T_{j})$ for trees in the $\tau$ ensemble. The complexity of the $\tau$ component of the model can be further reduced by using a smaller number of trees to estimate $\tau$ than for $\mu$. As we expect the magnitude of $\tau$ to be relatively small in comparison to $\mu$, it is also usual set the prior for the scale of the terminal node parameters in a $\tau$ tree, $\sigma_{\tau}^2$, to be less than $\sigma_{\mu}^2$. Secondly, if it is known that only a subset of the variables in $X$ are responsible for moderating the effect of $Z$ on $y$, then it is possible to use a different set of covariates in $\mu()$ and $\tau()$. Finally, as $\tau$ is now an explicit part of the model, it is possible to make direct inference on the treatment effects with BCF, and this provides a more straightforward interpretation of the model.

The procedure for fitting a BART model only requires minor adjustments for BCF to work. The modifications required are shown in Algorithm~\ref{BCFAlgorithm}.

\RestyleAlgo{boxruled}
\begin{algorithm}[!ht]
 \KwData{Target variable $y$ (length $n$; standardised), feature matrix $X$ ($n$ rows and $d$ columns), treatment variable Z (length $n$; 1 for treatment; 0 for control)}
 \KwResult{Posterior list of trees $T$, values of $\sigma$, fitted values $\hat{\mu}$, fitted values $\hat{\tau}$}
 \textbf{Initialisation\;}
 	Set Hyper-parameter values of $\alpha_{\mu}$, $\beta_{\mu}$, $\sigma_{\mu}$, $\sigma_{\tau}$, $\alpha_{\tau}$, $\beta_{\tau}$, $\nu$, $\lambda$\;
	Set Number of $\mu$ trees $N_{\mu}$, Number of $\tau$ trees $N_{\tau}$, Number of iterations $N$\;
	Set Initial value $\sigma = 1$, and set all $\mu$ trees and $\tau$ trees to stumps with terminal node parameters set to 0\;
\For{iterations $i$ from 1 to $N$}{
	\For{$\mu$ trees $j$ from 1 to $N_{\mu}$}{
		Compute partial residuals $R_{\mu,j}$ from $y$ minus predictions of all trees except $\mu$ tree $j$\;
		Grow a new tree $T_{\mu,j}^{new}$ based on grow/prune/change/swap\;
		Accept/Reject newly proposed tree structure with Metropolis-Hastings step using $P(T_{\mu,j}|R_{\mu,j}, \sigma)\propto P(T_{\mu,j})P(R_{\mu,j}|T_{\mu,j}, \sigma)$\;
		Sample $\mu$ values using $P(M_{\mu,j}|T_{\mu,j}, R_{\mu,j}, \sigma)$\;
	}
	\For{$\tau$ trees $k$ from 1 to $N_{\tau}$}{
		Compute partial residuals $R_{\tau,k}$ from $y$ minus predictions of all trees except $\tau$ tree $k$\;
		Grow a new tree $T_{\tau,k}^{new}$ based on grow/prune/change/swap\;
		Accept/Reject newly proposed tree structure with Metropolis-Hastings step using $P(T_{\tau,k}|R_{\tau,k}, \sigma)\propto P(T_{\tau,k})P(R_{\tau,k}|T_{\tau,k}, \sigma)$\;
		Sample $\tau$ values using $P(M_{\tau,k}|T_{\tau,k}, R_{\tau,k}, \sigma)$\;
	}
	Combine predictions from all trees to get $\hat{y}=\hat{\mu}+Z\hat{\tau}$\;
	Update $\sigma$ using $P(\sigma | \hat{y})$\;
}
\caption{\label{BCFAlgorithm}Bayesian Backfitting MCMC Algorithm for BCF}
\end{algorithm}

\subsection{Multivariate Bayesian Causal Forests}
\label{MVBCF Description}

Motivated by data from TIMSS which includes information on student achievement in both mathematics and science, we now extend the BCF algorithm to the multivariate setting. This extension allows us to estimate the causal effect of a given intervention on two or more outcomes jointly, and thus we are able to improve our predictions by taking advantage of the correlation between, and the shared information across all outcome variables. With our new setup, the BCF model specification becomes:

\[ \boldsymbol{Y}_{i}=\boldsymbol{\mu}_{i}+\boldsymbol{\tau}_{i}\circ\boldsymbol{Z}_{i} + \boldsymbol{\epsilon}_{i}\]
\noindent
where $\boldsymbol{Y}_{i}$ is a length $p$ vector representing the $i^{th}$ observation of the $p$ dimensional outcome variable $\boldsymbol{Y}$, $\boldsymbol{\mu}_{i}$ and $\boldsymbol{\tau}_{i}$ represent the $i^{th}$ predictions from the $\boldsymbol{\mu}(x_{i})$ and $\boldsymbol{\tau}(x_{i})$ functions, $\boldsymbol{\epsilon}_{i}$ is the $i^{th}$ residual, and $\circ$ is the Hadamard product operator. Note that with this setup we may allow the individual components of $\boldsymbol{Z}_{i}$, which indicate if a given treatment applies to those components of the outcome variable, to be different. This is an important feature because it allows us to apply our model to situations where the treatment or intervention may apply to the first dimension of the outcome $\boldsymbol{Y}$ (e.g. mathematics achievement), but not to the second dimension (e.g. science achievement) or vice-versa.

The MCMC algorithm for obtaining the posterior samples shares similarities with that of the univariate BCF model, but there are also a number of important differences. The tree prior is unchanged, as we allow all outcome variables to share the same tree structure. This is made appropriate by our motivating dataset, as we expect our chosen covariates will predict both outcomes in a similar way. As a result, the algorithm is encouraged to prioritise decision rules that will contribute positively towards accurately estimating all components of $\boldsymbol{Y}$. This helps to avoid over-fitting and acts as a type of regularisation, improving predictive performance. For the terminal node parameters we now place a multivariate normal prior over both $\boldsymbol{\mu}$ and $\boldsymbol{\tau}$:

\[\boldsymbol{\mu}_{j,k} \sim MVN \left( \boldsymbol{0} , \boldsymbol{\Sigma}_{\mu} = \sigma_{\mu}^2\boldsymbol{I} \right),\ \ \boldsymbol{\tau}_{j,k} \sim MVN \left( \boldsymbol{0} , \boldsymbol{\Sigma}_{\tau} = \sigma_{\tau}^2\boldsymbol{I} \right) \]
\noindent
Lastly, the conjugate prior for the residual covariance matrix is now an Inverse-Wishart distribution:

\[\boldsymbol{\Sigma} \sim \mathcal{W}^{-1}\left(\nu, \boldsymbol{\Sigma_{0}}\right)\]

\noindent
The extension to multivariate BCF requires substantially different updates to many of the parameters. For brevity we have listed these in the supplementary material rather than in the main text.

\section{Simulation Studies}
\label{Simulation Section}

In this section we present evidence of the advantages and improved predictive performance of the new multivariate approach. We do this by sharing the results of a simulation study in which we have compared the performance of our multivariate implementation of BCF with a univariate BCF model and a univariate S-Learner approach using BART from \citet{hill2011bayesian}. We compare out of sample model performance on three different target estimands:

\begin{enumerate}
    \item Predicted outcome under control; $\hat{y_{i}}(Z=0)$
    \item Predicted outcome under observed treatment status; $\hat{y_{i}}$
    \item Individual treatment effect estimates; $\hat{\tau_{i}}$
\end{enumerate}
\noindent
We also compare coverage and credible interval widths for all models tested.

Our simulated data come from a modified version of the first Friedman dataset \citep{friedman1991multivariate}, which is a commonly used benchmarking dataset within the machine learning literature as it provides a complicated non-linear pattern with complex interactions. The functional form of the first Friedman dataset is:

\[f(x_{i})=10 \sin(\pi x_{1,i}x_{2,i}) + 20(x_{3,i}-0.5)^2 + 10x_{4,i} + 5x_{5,i}\]
\noindent
where $x_{1}$, $x_{2}$, $x_{3}$, $x_{4}$, and $x_{5}$ are standard uniformly distributed random variables, and there are an additional five random variables $x_{6}$ to $x_{10}$, also uniformly distributed, which do not influence $y$ and serve as distractors.

In our multivariate causal tests we will create two outcome variables $Y_{1}$ and $Y_{2}$ which both use $f(x)$ as the true underlying pattern for $\mu$, the prognostic effect which we will interpret as the outcome under control. To decide which observations receive treatment, we will create two random variables $Z_{1}$ and $Z_{2}$, where $P(Z_{1}=1) \propto \mu$, and $P(Z_{2}=1) \propto -\mu$. We then consider two cases; one in which the effect of treatment is homogeneous, and one in which the effect is heterogeneous. In the homogeneous case we have that

\begin{align*}
\begin{pmatrix}
y_{i,1}\\
y_{i,2}
\end{pmatrix}
=
\begin{pmatrix}
\mu(x_{i}, \hat{\pi}_{i})\\
\mu(x_{i}, \hat{\pi}_{i})
\end{pmatrix}
+
\begin{pmatrix}
\tau_{1}\\
\tau_{2}
\end{pmatrix}
\circ
\begin{pmatrix}
Z_{i,1}\\
Z_{i,2}
\end{pmatrix}
+
\begin{pmatrix}
\epsilon_{i,1}\\
\epsilon_{i,2}
\end{pmatrix}
\end{align*}
\noindent
where $\epsilon_{i,1}$ and $\epsilon_{i,2}$ come from a multivariate normal distribution with mean $0$, covariance matrix $\sigma^{2}I_{2}$, and $\sigma^{2}$ is randomly generated with a single constraint that the signal to noise ratio must be between 2:1 and 1:1. The homogeneous treatment effects $\tau_{1}$ and $\tau_{2}$ are both randomly generated with magnitude less than $0.3$ times the standard deviation of $y$, as we a priori expect the effect of most treatments to be relatively small in comparison to the overall variance in $y$ \citep{kraft2020interpreting}, and it is important verify that multivariate BCF is capable of accurately detecting even a small effect size.

In the heterogeneous case, where the effect of treatment may be moderated by one or more of the observed covariates, we simulate

\begin{align*}
\begin{pmatrix}
y_{i,1}\\
y_{i,2}
\end{pmatrix}
=
\begin{pmatrix}
\mu(x_{i}, \hat{\pi}_{i})\\
\mu(x_{i}, \hat{\pi}_{i})
\end{pmatrix}
+
\begin{pmatrix}
\dfrac{1 + x_{i,6} + x_{i,7}}{2}\tau_{1}\\
\dfrac{2 + x_{i,7}}{3}\tau_{2}
\end{pmatrix}
\circ
\begin{pmatrix}
Z_{i,1}\\
Z_{i,2}
\end{pmatrix}
+
\begin{pmatrix}
\epsilon_{i,1}\\
\epsilon_{i,2}
\end{pmatrix}
\end{align*}
\noindent
with $\tau_{1}$ and $\tau_{2}$ as above, but the effect is now proportional to the moderating effect from $x_{6}$ and $x_{7}$ which play no role in generating $y$.

In the results that follow, we have generated 1000 synthetic data sets for the homogeneous test, and 1000 synthetic data sets for the heterogeneous test. Each dataset consists of 500 training observations, and 500 test observations per simulation. For multivariate and univariate BCF we have used 50 trees in the ensemble for estimating the prognostic effect $\mu$, and 20 trees in the ensemble for estimating the treatment effect $\tau$. For the BART approach we have used a total of 70 trees to estimate $y$. A total of 500 iterations were run for both the pre and post burn-in stages of model fitting. In each simulation we have fitted the multivariate BCF model to both outcome variables $Y_{1}$ and $Y_{2}$, and we have fitted all other approaches to outcome variables $Y_{1}$ and $Y_{2}$ separately. The R package BCF \citep{hahn2020bayesian} was used for the univariate implementation of BCF and the R package bartCause was used for the BART approach \citep{dorie2020package}. Default hyperparameter settings recommended by the authors were used in both cases.

\subsection{Results}

\begin{figure}
    \centering
    \includegraphics[height=15cm, angle=90]{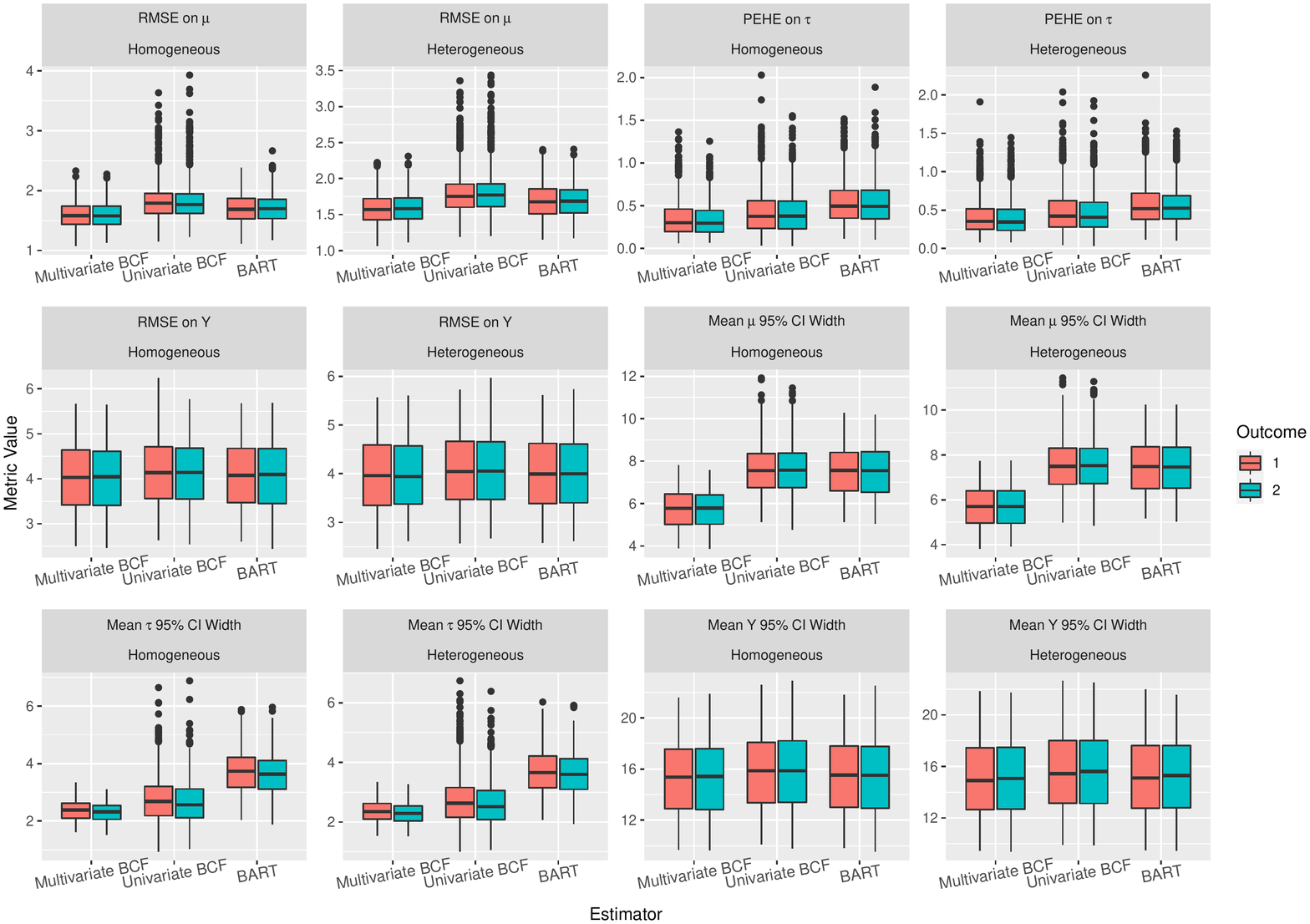}
    \caption{\label{sim results}Simulation Study Results}
\end{figure}

Figure~\ref{sim results} provides a graphical illustration of how the investigated approaches compare when predicting the outcome under control $\mu_{i}$, the individual treatment effects $\tau_{i}$, and the observed outcome $y_{i}$. We use the precision in estimating heterogeneous effects (PEHE; equivalent to the root mean squared error in estimating $\tau$) to evaluate predictive performance when estimating $\tau_{i}$: $PEHE=\sqrt{\frac{1}{N}\sum_{i=1}^{N}(\tau_{i}-\hat{\tau_{i}})^2}$. A numerical summary of our results can also be found in Table~\ref{simulation table}.


\begin{table}
\begin{center}
\rowcolors{5}{}{black!10}
\begin{tabular}{l p{0.55cm}p{0.55cm}p{0.55cm}p{0.55cm}p{0.55cm}p{0.55cm}p{0.55cm}p{0.55cm}p{0.55cm}p{0.55cm}p{0.55cm}p{0.55cm}}
\toprule
 & \multicolumn{6}{l}{Homogeneous Treatment Effect} & \multicolumn{6}{l}{Heterogeneous Treatment Effect} \\
\cmidrule(lr){2-7} \cmidrule(lr){8-13}
 & \multicolumn{2}{l}{MVBCF} & \multicolumn{2}{l}{BCF} & \multicolumn{2}{l}{BART} & \multicolumn{2}{l}{MVBCF} & \multicolumn{2}{l}{BCF} & \multicolumn{2}{l}{BART} \\
 \cmidrule(lr){2-3} \cmidrule(lr){4-5} \cmidrule(lr){6-7} \cmidrule(lr){8-9} \cmidrule(lr){10-11} \cmidrule(lr){12-13}
 \rowcolor{gray!0}
Metric            &$Y_{1}$&$Y_{2}$&$Y_{1}$&$Y_{2}$&$Y_{1}$&$Y_{2}$&$Y_{1}$&$Y_{2}$&$Y_{1}$&$Y_{2}$&$Y_{1}$&$Y_{2}$ \\
\midrule
RMSE on $\mu$             & \textbf{1.58} & \textbf{1.58} & 1.79 & 1.81 & 1.69 & 1.69 & \textbf{1.60} & \textbf{1.60} & 1.80 & 1.80 & 1.71 & 1.70 \\
PEHE on $\tau$            & \textbf{0.34} & \textbf{0.34} & 0.41 & 0.40 & 0.52 & 0.52 & \textbf{0.41} & \textbf{0.41} & 0.49 & 0.48 & 0.57 & 0.58 \\
RMSE on $y$               & \textbf{3.98} & \textbf{3.96} & 4.07 & 4.07 & 4.02 & 4.01 & \textbf{4.01} & \textbf{4.02} & 4.10 & 4.10 & 4.05 & 4.06 \\
CRPS on $\mu$             & \textbf{0.89} & \textbf{0.89} & 1.02 & 1.02 & 0.96 & 0.96 & \textbf{0.89} & \textbf{0.89} & 1.01 & 1.02 & 0.96 & 0.95 \\
CRPS on $\tau$            & \textbf{0.24} & \textbf{0.23} & 0.28 & 0.27 & 0.34 & 0.34 & \textbf{0.27} & \textbf{0.26} & 0.31 & 0.29 & 0.36 & 0.35 \\
CRPS on $y$               & \textbf{2.27} & \textbf{2.27} & 2.33 & 2.32 & 2.30 & 2.29 & \textbf{2.24} & \textbf{2.25} & 2.30 & 2.30 & 2.27 & 2.27 \\
\bottomrule
\end{tabular}
\caption{\label{simulation table} Simulation study results for $y_{1}$ and $y_{2}$. Best results highlighted in bold.}
\end{center}
\end{table}


As can be seen from Figure~\ref{sim results}, multivariate BCF performs favourably in comparison to both univariate BCF and the BART approach when predicting $\mu$, the outcome under control. This improvement in performance can be attributed to the ability of multivariate BCF to jointly consider multiple outcome variables, thus allowing the predictions made by the model to benefit from its awareness of the shared variance across these outcomes. This improvement is seen in the results for both the homogeneous and the heterogeneous simulation study.

When predicting $\tau$ (the individual level treatment effects) we can again see the advantages of the multivariate approach, as multivariate BCF outperforms univariate BCF and the BART approach in this test as well. This is to be expected, because if a better estimate for $\mu$ can be obtained, then the estimates for $\tau$ which are directly linked to $\mu$ should benefit from this increase in accuracy. Note that this performance was achieved even when we allowed the treatment variables $Z_{i,1}$ and $Z_{i,2}$ to differ. This added layer of flexibility allows our model to be applied to situations where an intervention may be directed at one component of the outcome variable $Y$, but not the others, as may be the case if a student attends extra lessons lessons for mathematics class but not for science class or vice-versa. 

All three methods perform approximately equally as well when assessed on their performance predicting $y$, the outcome under a unit's observed treatment status. This is pleasantly surprising, since it demonstrates that the two BCF approaches are competitive in a situation (predicting the outcome $y$) which is more suited to standard BART. It is a good illustration of the impressive performance of BCF as it is able to accurately estimate two separate components of $y$, thus improving interpretation, without sacrificing any accuracy. 

When we look at the 95\% credible interval widths for $\mu$, $\tau$ and $y$ we see that the widths are narrower for multivariate BCF than for the univariate BCF and the BART approach. This implies that multivariate BCF has greater confidence in its predictions than the other approaches, and coincides with the greater predictive accuracy demonstrated by the model. This is true for all three of the target estimands: $\mu$, $\tau$, $y$, and across both simulation types.

Table~\ref{simulation table} also provides the mean Continuous Ranked Probability Score \citep[(CRPS)][]{matheson1976scoring} metric values for each estimand and simulation type. The CRPS takes into account the full posterior distribution of each prediction and thus provides a good indication of how well the posterior distributions are calibrated. Lower scores are better for CRPS, and multivariate BCF once again outperforms the two univariate approaches, demonstrating not only superior predictive performance, but also well calibrated posterior distributions for the $\mu$, $\tau$ and $y$ estimates.

To summarise, the results from this section have demonstrated the practical benefits of employing a multivariate approach when estimating the causal effect of an intervention on two correlated outcomes. This is evident from the improved performance in multivariate BCF, which outperformed its univariate equivalent in all three tasks: predicting $y$, $\mu$, and the treatment effect $\tau$. This was true across both simulated datasets: one with a homogeneous treatment effect, and one with a heterogeneous treatment effect moderated by two covariates. Good coverage was also achieved in each test. Encouraged by the impressive performance of multivariate BCF, we now proceed to apply our model to a real dataset from the world of education in the next section.

\section{Application to TIMSS 2019}
\label{Application Section}

In this section, we describe the data used from TIMSS 2019 before applying our multivariate BCF model to investigate the effect of three different treatments on student achievement in mathematics and science.

\subsection{Data Description and Procedure}

TIMSS 2019 is the seventh cycle of TIMSS to have taken place, with a total of 64 countries participating across the fourth and eighth grade components of the study, making TIMSS 2019 one of largest installments of the programme to date \citep{mullis2020timss}. For the purposes of this study however, we will restrict our attention to the eighth grade subset of the data from Ireland. This subset of the data provides us with a representative sample of 4118 eighth grade secondary school students; 2118 male, 1948 female, and 52 who did not say with an average age of 14.42 years. In addition to this, the mathematics and science teachers of these students participated in the study, providing us with data on 565 mathematics teachers and 409 science teachers. The students' school principals participated too, giving us a total of 149 principal questionnaire observations. 

After merging each student's data together with that of their mathematics teacher, science teacher, and school principal, the end product is a combined dataset of 4118 observations, each comprising 50 variables describing various student, teacher, and school characteristics. Important student level characteristics include gender, age, attitude towards and motivation for studying mathematics and science, as well as how many books are in their home, and the highest level educational qualification received by both parents. Important teacher level characteristics include number of years' teaching experience, area of study during their degree, perceptions of the school's level of emphasis on academic success, and teaching practices within the classroom. From the principal data we also have access to information such as the number of students in the school, a description of the average socioeconomic background of the students of the school, and a summary of how well resourced the school is in general. We will control for these variables as potential confounders as we investigate the three factors described in Section~\ref{TIMSS Section}. A complete list of all variables used can be found in the supplementary material.

TIMSS 2019 used a stratified two-stage cluster sample design to ensure that the data gathered can be used as a nationally representative sample of the population of eighth grade students within a country \citep{martin2020methods}. As part of this complex survey design, students taking part in the study are assigned a sampling weight to indicate how many students in the total population they are representative of. These weights were accounted for in our study by appropriately weighting the treatment effect estimates of individual students when calculating the average treatment effect for the student population.

One extra complicating factor that must be addressed when working with data from TIMSS is the use of plausible values for student achievement. It is difficult for TIMSS to accurately estimate student achievement with only a limited number of mathematics/science questions, and this is further complicated by the fact that not all students answer the same booklet of questions during the TIMSS study. Therefore, instead of providing a single achievement estimate of each student, the TIMSS study organisers have drawn five plausible values from the posterior distribution of each student's achievement level in order to better represent the underlying uncertainty that is present. These five plausible values which are available in the public data were fully accounted for in our study. Five MCMC chains were run for every model, each corresponding to one of the five plausible values, and these chains were pooled together after burn-in.

To obtain the results in the following section, each plausible value chain was run for 3500 burn-in and 1500 post burn-in iterations. A total of 50 BART trees were used for the prognostic component $\mu()$, while a smaller number of 20 BART trees were used for estimating the treatment effects $\tau()$. Satisfactory convergence was assessed via visual inspection of samples from the variance covariance matrix $\Sigma$, predicted values $\hat{\mu}$ and $\hat{\tau}$ for a random sample of individuals in the dataset, and the ATE estimates themselves.

\subsection{Results}

\begin{figure}
    \centering
    \includegraphics[width=14cm]{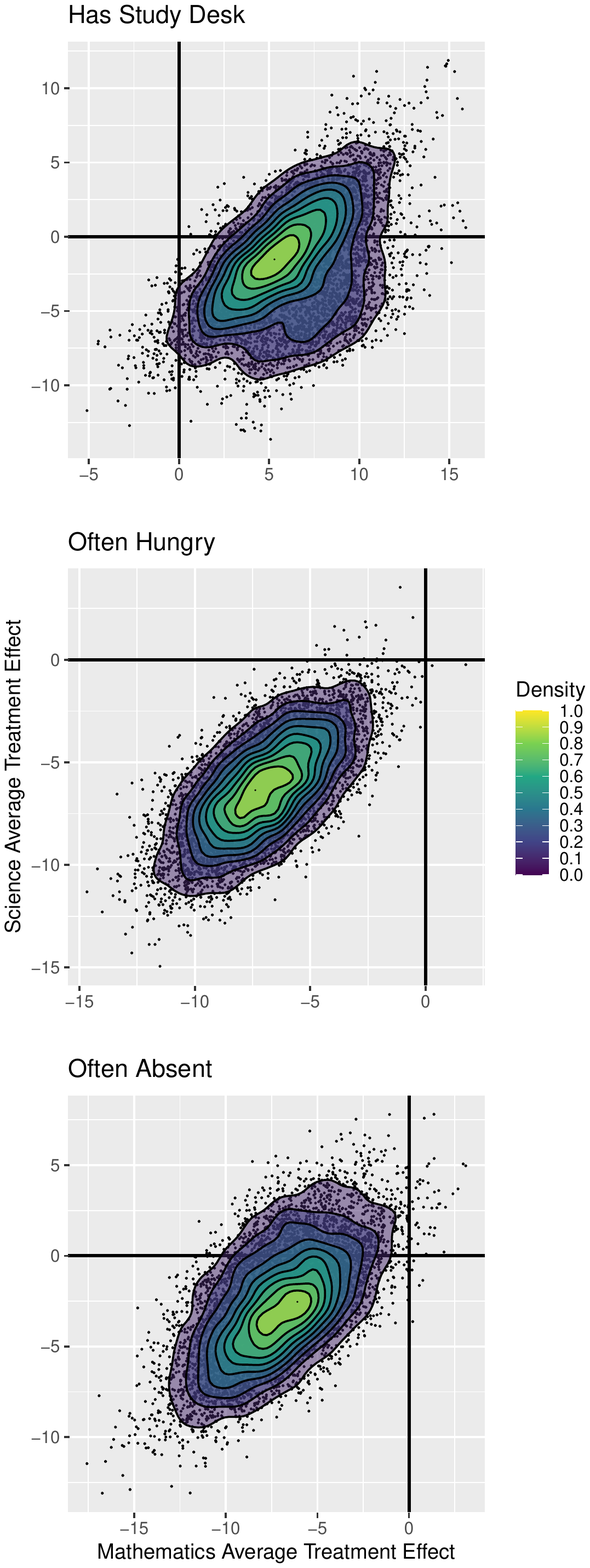}
    \caption{\label{ate results}Plot of average treatment effect results. Each section displays a density plot of the sampled posterior distribution of the average treatment effects for mathematics ($x$-axis) and science ($y$-axis).}
\end{figure}

Figure~\ref{ate results} shows a density plot of the posterior distributions of the average treatment effects for each of the treatments we have investigated. Each point in a scatter plot corresponds to the weighted average of the individual treatment effect estimates from a single iteration of one of the five plausible value MCMC chains. Credible intervals for all of these treatments can be found in Table~\ref{timss table} which also provides the control and treatment group sizes for each intervention under investigation. The treatment group size for "Has Study Desk" is 3672, indicating that 89\% of the students in the sample did report having a study desk at home, while the remaining 11\% did not. The control and treatment group sizes for often being hungry when arriving at school or often being absent have similar interpretations. A broader discussion of the wider context of these results can be found in the following section, but for now we will focus only on a summary.

\begin{table}
\begin{center}
\begin{tabular}{ p{3cm}p{3cm}p{3cm}p{3cm}  }
\toprule
 \multicolumn{4}{c}{TIMSS Results} \\
 \hline
  & Has Study Desk & Often Hungry at School & Often Absent\\
 \hline
 Treatment Group Size   & 3672 (89\%)    &954 (23\%)&   503 (12\%)\\
 Control Group Size&   446 (11\%)  & 3164 (77\%)   &3615 (88\%)\\
 Mathematics ATE & 6.10 & -6.98&  -7.08\\
 Science ATE & -1.93 & -6.24&  -2.74\\
 Mathematics 95\% CI&  (0.20, 11.67)  & (-11.15, -2.78)& (-12.47, -1.55)\\
 Science 95\% CI& (-8.57, 5.34)  & (-10.82, -1.72)   & (-8.44, 3.42)\\
 \hline
\end{tabular}
\caption{\label{timss table} Application to TIMSS Data Results}
\end{center}
\end{table}

To assist in the interpretation of the results that follow, consider that student achievement at the eighth grade in mathematics in Ireland is approximately normally distributed with mean 524, standard deviation 73, and students at the 10th and 90th percentiles scoring approximately 432 and 614 respectively. Therefore, a treatment effect magnitude of 7.3 would correspond to a 0.1 standard deviation increase/decrease in student achievement in mathematics. Effect sizes of this magnitude are common in educational studies and can be thought of as being "medium" in size \citep{kraft2020interpreting}. Science achievement follows a very similar distribution with mean 523 and standard deviation 83.

Multivariate BCF clearly identifies access to a study desk at home as having a positive impact on student achievement in mathematics. The effect of having a desk on student achievement in science is less clear however, and the average treatment effect is centered very close to 0. Our results for the second treatment under investigation, often being hungry at school, show that this factor is associated with a very negative impact on both mathematics and science achievement. The magnitude of the effect identified is almost identical for both mathematics and science achievement. Finally, often being absent from school is also identified as having a negative impact on achievement in both mathematics and science. The impact on science achievement however, as was the case with having access to a desk, is slightly less clear than for mathematics achievement. The posterior distributions of the average treatment effects for mathematics and science achievement are positively correlated for all three treatments under investigation. This agrees with our intuition that the effect any of these three factors may have is likely to be similar on achievement in both subjects. We did observe some differences in magnitude however, most notably in relation to the "Has Study Desk" treatment.

\begin{figure}[!h]
    \centering
    \includegraphics[width=14cm]{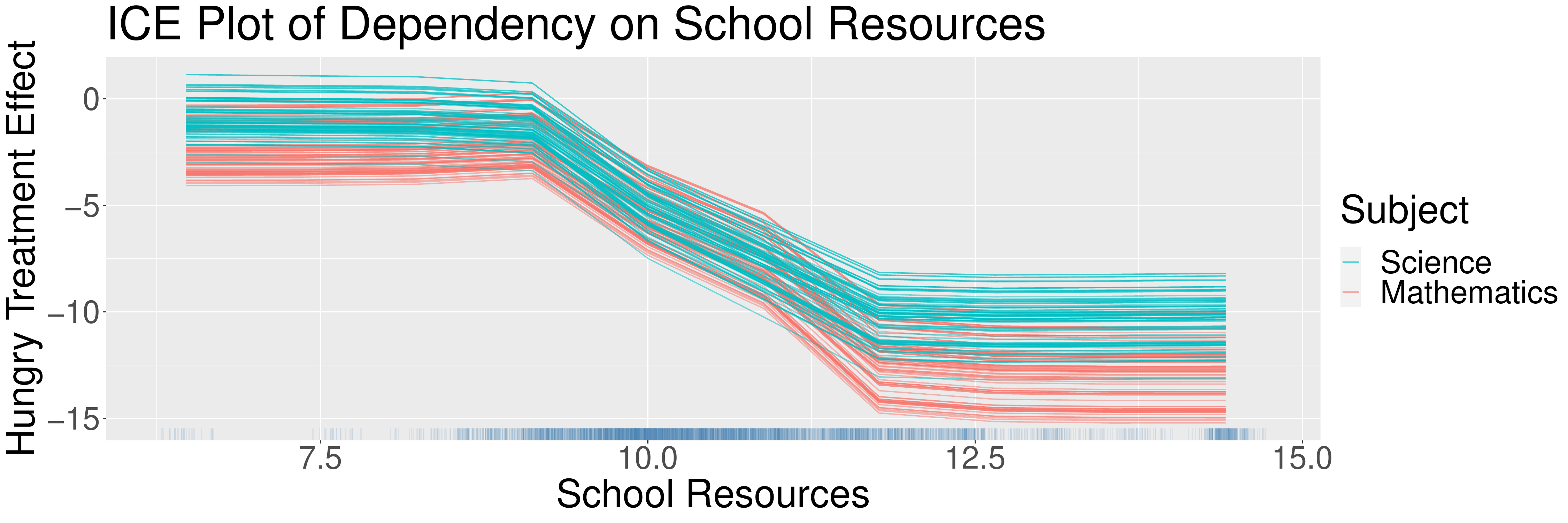}
    \caption{\label{hungryiceplot}ICE Plot of the moderating role of school resources on the "Often Hungry" treatment effect. (Random sample of 100 students to avoid overprinting). A jittered rug has been added to the $x$-axis to display the distribution of the average school resources variable. Students in schools with less resources appear to be less negatively affected by arriving to school feeling hungry.}
\end{figure}

\begin{figure}[!h]
    \centering
    \includegraphics[width=14cm]{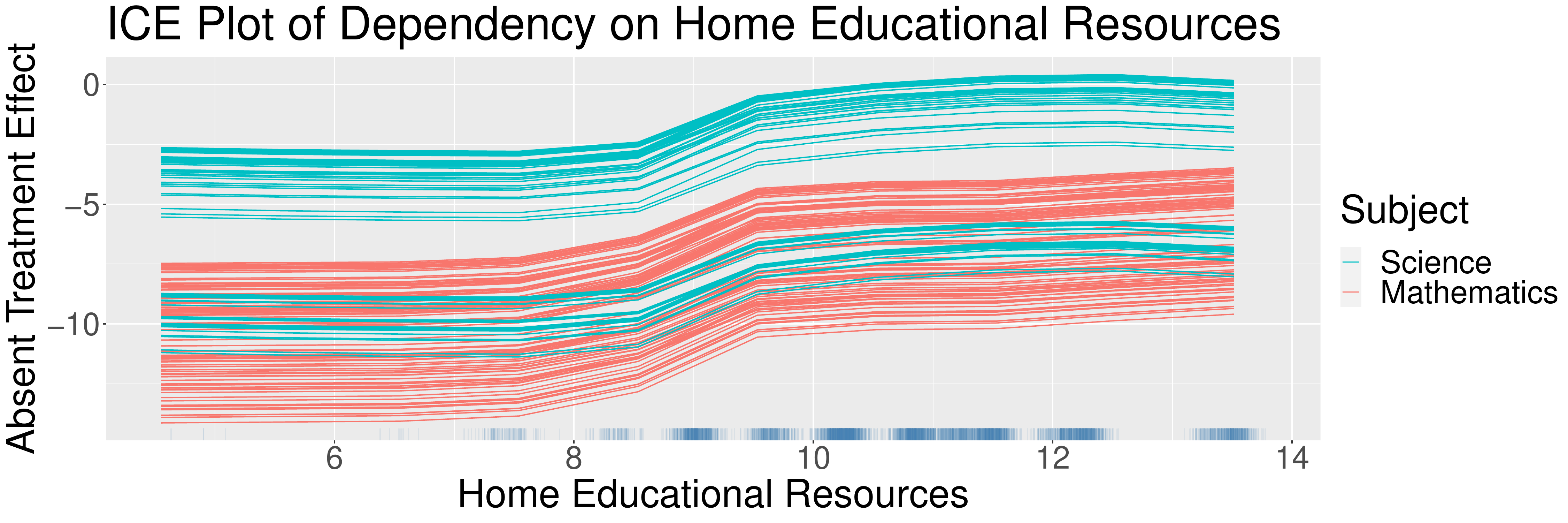}
    \caption{\label{absentresources}ICE Plot of the moderating role of home educational resources on the "Often Absent" treatment effect. (Random sample of 100 students to avoid overprinting). A jittered rug has been added to the $x$-axis to display the distribution of home resources resources variable. Students with more educational resources at home appear to be less negatively affected by regular absences. Notice the two clusters of blue lines which correspond to students who know and don't know their parent's education level (See Figure~\ref{absenteducation}).}
\end{figure}

\begin{figure}[!h]
    \centering
    \includegraphics[width=14cm]{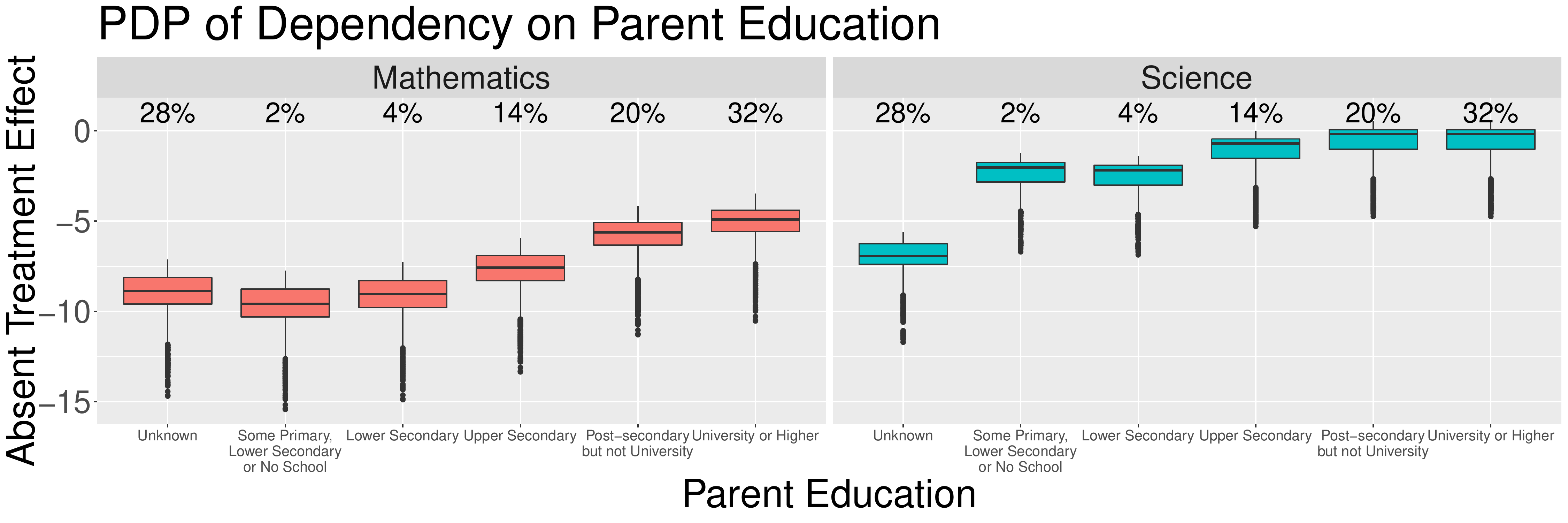}
    \caption{\label{absenteducation}PDP of the moderating role of parent education on the "Often Absent" treatment effect. Students with highly educated parents appear to be less negatively affected by regular absences. Percentages indicate proportion of students belonging to each category.}
\end{figure}

One disadvantage of the posterior distributions plotted in Figure~\ref{ate results} is that they only provide a summary of the average treatment effects. Often, however, the effects of a treatment felt by an individual may be modified by one or more moderating variables. To examine this possibility and to investigate the moderating effect of these variables we have created individual conditional expectation \citep[(ICE)][]{goldstein2015peeking} and partial dependence plots \citep[(PDP)][]{friedman2001greedy} of the treatment effects which visualise the dependency of the treatment effects on these covariates. Figure~\ref{hungryiceplot} shows the results for the treatment "Often Hungry" which exhibits an interesting trend, as it would appear students in schools with less resources tend to experience a less negative treatment effect. Schools in disadvantaged areas with fewer resources are more likely to receive access to free school meal programmes in Ireland \citep{schoolMeals}, so this is possibly an indication that free school meal programmes are successfully mitigating the negative consequences of students often arriving at school feeling hungry. Without knowing which schools do in fact participate in free school meal programmes, however, we can only speculate on the true moderating role of school resources here. A different pattern is observed in Figures~\ref{absentresources} and~\ref{absenteducation} which show that students with more educated parents and more home resources are less negatively affected by frequent absences from school. These students may be in a better position to "catch up" on missed school work due to the physical and parental resources available to them, but again we can only hypothesise regarding the true moderating role of home educational resources and parent education in relation to often missing school.

\section{Discussion}

In this paper, motivated by data from the Trends in International Mathematics and Science Study which includes information on student achievement in both mathematics and science, we have developed a multivariate extension of Bayesian Causal Forests which can be used to estimate the causal effect of an intervention on two or more outcome variables simultaneously. The key advantage of our approach is the use of the same tree structure for both outcome variables. This enables us to leverage the shared variance across the outcomes, resulting in improved predictive performance, as demonstrated in our simulation study. An interesting feature of our model is that it allows the treatment indicator $Z$ to be different for both outcomes. This is important when a treatment can be applied to the outcome variables independently, for example when a student can receive homework in one subject but not the other. This feature was not needed in our analysis of the TIMSS data, as the causal factors under investigation were not subject specific, but its utility was demonstrated during our simulation study.

The results from our simulation study indicate that the multivariate BCF model is capable of estimating causal effects with a greater level of accuracy than both the univariate BCF model and the BART S-Learner approach. Our multivariate model's performance was even comparable to that of BART when predicting $y$, which should excel at this task as this is its only objective. This increase in performance was observed across both the homogeneous and heterogeneous simulated data sets. As well as observing an increase in predictive performance, it was also noted that multivariate BCF was able to attain good posterior calibration with narrower credible intervals, indicating the model is capable of accurately estimating treatment effects with a higher degree of precision and certainty in its estimates.

In our application of the multivariate BCF model to the motivating TIMSS dataset we found that access to a study desk at home is associated with a clear increase in mathematics achievement, but with no discernible change in science achievement. Unsurprisingly, often being hungry at school was identified as having a very negative impact on achievement in both mathematics and science. A similar negative effect was found to be associated with often being absent from school. These results agree with findings from previous studies within the field of education which have identified the importance of home-related factors for predicting student achievement in mathematics and science \citep[e.g.][]{tsai2015hierarchical, vik2022aspects, vesic2021role}. This study therefore makes an important contribution by verifying these results within a causal machine learning framework. 

Our results provide further evidence of the potential for targeted interventions such as free school meal programmes to tackle the negative consequences of students frequently lacking a healthy breakfast in the morning, or a lunch while they are at school. This is made clear by the very negative effects of being hungry at school which such schemes may help to avoid. The positive effect of having a study desk may also indicate an opportunity to inform parents about the importance of students having dedicated study spaces at home. Finally, the clear negative impact that was observed from students often being absent may highlight the potential for schools to investigate these absences, and to prepare extra supports for affected students. 

One limitation of multivariate tree based models is that they can struggle when the outcome variables of interest are weakly or not at all correlated. In some settings it may be the case that a very different tree structure is appropriate for the outcome variables, and in these situations the requirement that both outcome variables share the same tree structure can be quite restrictive, leading to a reduction in overall performance. \citet{rahman2017integratedmrf} for example found that multivariate extensions of the random forest machine learning model performed better for highly correlated outcomes. It is likely that similar limitations would apply to our model as well. This is unlikely to apply to our investigation of the TIMSS data because there is a strong positive correlation between student achievement in mathematics and science, $r=0.85$, and it is likely that variables related to a student's family's socioeconomic status or school will have a similar effect on achievement in both subjects, thus making a shared tree structure very appropriate.

Potential avenues for further development of the multivariate BCF approach could include allowing a subset $\mathcal{M}_{1}$ of the BART trees in the $\mu()$ ensemble to independently predict the first outcome, while allowing a second subset $\mathcal{M}_{2}$ to predict the second outcome. This greater flexibility would allow multivariate BCF to more easily account for the possibly different tree structures which may be suited to highly uncorrelated outcome variables. A similar modification could also be applied to the ensemble of trees responsible for predicting $\tau$ if there is reason to believe that the treatment effect is likely to have a very different impact on both outcomes, and the effect is likely to be moderated by a distinct set of variables for each outcome. 

Finally, although our motivating dataset came from the world of education in this study, it is also likely that our multivariate approach would be useful in other fields such as economics or medicine. A researcher may be interested for example in the effect of a drug $D$ on both the systolic and diastolic blood pressure of patients who have been prescribed it by their doctor. Areas for future research therefore also include the application of multivariate BCF to other disciplines with multivariate outcomes of interest.

\section*{Disclosure Statement}

The authors declare no conflict of interest.

\section*{Funding}

This work has emanated from research conducted with the financial support of Science Foundation Ireland under grant number 18/CRT/6049. In addition Andrew Parnell's work was supported by: a Science Foundation Ireland Career Development Award (17/CDA/4695) and SFI Research Centre award (12/RC/2289\_P2). For the purpose of Open Access, the authors have applied a CC BY public copyright licence to any Author Accepted Manuscript version arising from this submission.

\bibliography{refs.bib}

\newpage
\appendix
\section{Multivariate BCF Updates}
\setcounter{page}{1}
\label{appendixA}

\subsection{Log-Likelihood of a $\mu$ tree}

Let $T_{j}$ denote the $j^{th}$ $\mu$ tree in the ensemble with partial residuals $R_{j}$. Also suppose that tree $T_{j}$ has $K$ terminal nodes $h_{1} ... h_{K}$, and $L$ non-terminal nodes $b_{1} ... b_{L}$. Furthermore, let $R_{k,1} ... R_{k,n_{k}}$ denote the partial residuals which fall into the $k^{th}$ terminal node of tree $T_{j}$. Then given the residual covariance matrix $\Sigma$, the tree priors $\alpha$ and $\beta$, and the prior covariance matrix $\Sigma_{\mu}$ for terminal node parameters, we have that:

\[\ell(T_{j}|R_{j}, \Sigma) \propto \ell(R_{j}|T_{j}, \Sigma) + \ell(T_{j})\]
\[\ell(T_{j})=\sum_{k=1}^{K} \log \left(1 - \alpha(1 + d(h_{k}))^{-\beta} \right) + \sum_{l=1}^L \log(\alpha) - \beta \log(1 + d(b_{l})) \]
\begin{align*}
\ell(R_{j}|T_{j}, \Sigma) &= \sum_{k=1}^K \ell(R_{k,1}, \ldots, R_{k,n_k} | T_{j}, \Sigma) \\
&\propto \sum_{k=1}^K \left\{-\dfrac{n_{k}}{2} \log\left(\left| \Sigma \right|\right) -\dfrac{1}{2} \log\left(\left| \Sigma_{\mu}\right|\right) + \dfrac{1}{2} \log\left(\left| \Sigma_{k,0}\right|\right) -\dfrac{1}{2}\sum ^{n_{k}}_{i=1}\left(R_{k,i}^{T}\Sigma^{-1}R_{k,i} - \mu_{k,0}^{T}\Sigma _{k,0}^{-1}\mu_{k,0}\right) \right\}
\end{align*}

where
\[\Sigma_{k,0}^{-1}=n_{k}\Sigma^{-1}+\Sigma_{\mu}^{-1}\]
and
\[\mu_{k,0}=\Sigma_{k,0}\Sigma^{-1}\left(\sum ^{n_{k}}_{i=1}R_{k,i}\right)\]

\subsection{Posterior distribution of terminal node parameters in a $\mu$ tree}

For the $k^{th}$ terminal node of any tree $T_{j}$, the posterior distribution of the terminal node parameter $\mu_{j,k}$ with prior mean $\mu_{0}$ is given by:

\[\mu_{k}| \ldots \sim N \left( \mu_{n} , \Sigma_{n} \right)\]

where

\[\mu_{n}=\left( \Sigma_{\mu}^{-1}+n_{k}\Sigma^{-1}\right)^{-1}\left(\Sigma_{\mu}^{-1}\mu_{0}+n_{k}\Sigma^{-1}\overline{R}\right)\]

and

\[\Sigma_{n}=\left( \Sigma_{\mu}^{-1}+n\Sigma^{-1}\right)^{-1}\]

\subsection{Log-Likelihood of a $\tau$ tree}

Given the prior covariance matrix for terminal node parameters $\Sigma_{\tau}$, and $k \times p$ matrix $Z_{k}$ which holds the treatment status of each observation in terminal node $k$, who's transposed $i^{th}$ row we denote by $Z_{k,i}$, we obtain:

\[\ell(T_{j}|R_{j}, \Sigma) \propto \ell(R_{j}|T_{j}, \Sigma) + \ell(T_{j})\]
\[\ell(T_{j})=\sum_{k=1}^{K} \log \left(1 - \alpha(1 + d(h_{k}))^{-\beta} \right) + \sum_{l=1}^L \log(\alpha) - \beta \log(1 + d(b_{l})) \]

\begin{align*}
\ell(R_{j}|T_{j}, \Sigma) &= \sum_{k=1}^K \ell(R_{k,1}, \ldots, R_{k,n_k} | T_{j}, \Sigma) \\
&\propto \sum_{k=1}^K \left\{-\dfrac{n_{k}}{2} \log\left(\left| \Sigma \right|\right) -\dfrac{1}{2} \log\left(\left| \Sigma_{\tau}\right|\right) + \dfrac{1}{2} \log\left(\left| \Sigma_{k,0}\right|\right) -\dfrac{1}{2}\sum ^{n_{k}}_{i=1}\left(R_{k,i}^{T}\Sigma^{-1}R_{k,i} - \tau_{k,0}^{T}\Sigma _{k,0}^{-1}\tau_{k,0}\right) \right\}
\end{align*}

where
\[\Sigma_{k,0}^{-1}=Z_{k}^TZ_{k}\Sigma^{-1}+\Sigma_{\tau}^{-1}\]

\[\tau_{k,0}=\Sigma_{k,0}\sum^{n_{k}}_{i=1}Z_{k,i} \circ \Sigma ^{-1}R_{k,i}\]

\subsection{Posterior distribution of terminal node parameters in a $\tau$ tree}

Analogously to the terminal node parameters in a $\mu$ tree, the posterior distribution of the terminal node parameter $\tau_{j,k}$ in the $k^{th}$ leaf of the $j^{th}$ $\tau$ tree $T_{j}$, with prior mean $\tau_{0}$ is: 

\[\tau_{j,k}| \ldots \sim N \left( \tau_{n} , \Sigma_{n} \right)\]

where

\[\tau_{n}=\left(\Sigma_{\tau}^{-1}+Z_{k}^{T}Z_{k}\circ\Sigma^{-1}\right)^{-1}\left(\Sigma_{\tau}^{-1}\tau_{0}+\sum_{i=1}^{n_{k}}Z_{k,i}\circ[\Sigma^{-1}R_{k,i}]\right)\]

and

\[\Sigma_{n}=\left( \Sigma_{\tau}^{-1}+Z_{k}^{T}Z_{k}\circ\Sigma^{-1}\right)^{-1}\]

\subsection{Posterior distribution of residual covariance parameter $\Sigma$}

Given observed and predicted values $y$ and $\hat{y}$, the posterior distribution of the covariance matrix $\Sigma$ for the $n$ residuals, with prior scale matrix $\Sigma_{0}$ and $\nu_{0}$ degrees of freedom is given by:

\[\Sigma| \ldots \sim inv-Wish \left( \nu_{0}+n , [ S_{0}+S_{\theta}]^{-1} \right)\]

where 

\[S_{\theta}=\sum_{i=1}^n(y_{i}-\hat{y_{i}})^T\Sigma^{-1}(y_{i}-\hat{y_{i}}).\]

\newpage
\section{TIMSS Variables Used in Study}
\label{AppendixB}

\begin{table}[h!]
\scriptsize
\begin{center}
\begin{tabular}{ p{3cm}p{3cm}p{7cm}  }
\toprule
 \multicolumn{3}{c}{Variables Used} \\
 \hline
 Variable Code & Obtained From  & Description\\
 \hline
 BSDAGE   & Student Questionnaire    & Student Age\\
 BSBG01&   Student Questionnaire  & Student Gender   \\
 BSBG03 & Student Questionnaire & How often student speaks English at home\\
 BSBG04 & Student Questionnaire & Number of books at home\\
 BSBG07& Student Questionnaire & How far in education student expects to go\\
 BSBG08A& Student Questionnaire  & Was parent/guardian A born in Ireland\\
 BSBG08B& Student Questionnaire  & Was parent/guardian B born in Ireland\\
 BSBG09A& Student Questionnaire  & Was student born in Ireland\\
 BSBG10&  Student Questionnaire & How often student is absent\\
 BSBG11A& Student Questionnaire  & How often student feels hungry when arriving at school\\
 BSBG11B&  Student Questionnaire & How often student feels tired when arriving at school\\
 BSDGEDUP& Student Questionnaire  & Parent's highest education level\\
 BSBGHER& Student Questionnaire  & Number of home educational resources\\
 BSBGSSB& Student Questionnaire  & Sense of school belonging\\
 BSBGSB& Student Questionnaire  & School bullying\\
 BSBGSCM/BSBGSCS & Student Questionnaire  & Confidence in mathematics/science\\
 BSBGSVM/BSBGSVS & Student Questionnaire  & Student values mathematics/science\\
 BSBGICM/BSBGICS& Student Questionnaire  & Instructional clarity in mathematics/science\\
 BSBG05A& Student Questionnaire  & Has computer/tablet at home\\
 BSBG05B& Student Questionnaire  & Has study desk at home\\
 BSBG05C& Student Questionnaire  & Has own bedroom\\
 BSBG05D& Student Questionnaire  & Has home internet connection\\
 BSBG05E& Student Questionnaire  & Has own mobile phone\\
 BSBG05F& Student Questionnaire  & Has gaming system\\
 BSBG05G& Student Questionnaire  & Home TV has "premium" TV channels\\
 BTBG01& Teacher Questionnaire  & Number of years teaching\\
 BTBG02&  Teacher Questionnaire & Teacher gender\\
 BTBG03& Teacher Questionnaire  & Teacher age\\
 BTBG10& Teacher Questionnaire  & Number of students in class\\
 BTBGTJS& Teacher Questionnaire  & Teacher job satisfaction\\
 BTBGSOS& Teacher Questionnaire  & Safe and orderly school\\
 BTBGLSN& Teacher Questionnaire  & Teaching is limited by students not ready for instruction\\
 BTBGEAS& Teacher Questionnaire  & Emphasis on academic success\\
 BTDMME& Teacher Questionnaire  & Type of degree\\
 BCBGDAS& Principal Questionnaire  & School discipline\\
 BCBGEAS& Principal Questionnaire  & Emphasis on academic success\\
 BCBGMRS/BCBGSRS& Principal Questionnaire  & Resource shortages in mathematics/science\\
 BCDGSBC& Principal Questionnaire  & School average socioeconomic background\\
 
 \hline
\end{tabular}
\caption{\label{variable table} Control Variables Used}
\end{center}
\end{table}

\end{document}